\documentclass[letterpaper]{article} 
\usepackage[]{arXiv}  
\usepackage{times}  
\usepackage{helvet}  
\usepackage{courier}  
\usepackage[hyphens]{url}  
\usepackage{graphicx} 
\usepackage{natbib}  
\usepackage{caption} 
\nocopyright

\usepackage{amssymb}
\usepackage{amsmath}
\usepackage{algorithm}
\usepackage{algpseudocode}
\usepackage{caption}
\usepackage{subcaption}
\usepackage{bbm}
\usepackage{multirow}
\usepackage[table]{xcolor}

\usepackage{newfloat}
\usepackage{listings}
\DeclareCaptionStyle{ruled}{labelfont=normalfont,labelsep=colon,strut=off} 
\floatstyle{ruled}
\newfloat{listing}{tb}{lst}{}
\floatname{listing}{Listing}
\newcommand{\dataset}[1]{{\tt #1}}
\newcommand{\algo}[1]{{\tt #1}}

\title{Improving Model Training via Self-learned Label Representations}
\author {
	Xiao Yu,
	Nakul Verma
}
\affiliations {
    Columbia University\\
	xy2437@columbia.edu, 
	verma@cs.columbia.edu
}

\begin{document}

\maketitle

\begin{abstract}
Modern neural network architectures  have shown remarkable success in several large-scale classification and prediction tasks. Part of the success of these architectures is their flexibility to transform the data from the raw input representations (e.g.\ pixels for vision tasks, or text for natural language processing tasks) to one-hot output encoding. While much of the work has focused on studying how the input gets transformed to the one-hot encoding, very little work has examined the effectiveness of these one-hot labels. 

In this work, we demonstrate that more sophisticated label representations are better for classification than the usual one-hot encoding. We propose Learning with Adaptive Labels (\algo{LwAL}) algorithm, which simultaneously learns the label representation while training for the classification task. These learned labels can significantly cut down on the training time (usually by more than 50\%) while often achieving better test accuracies. Our algorithm introduces negligible additional parameters and has a minimal computational overhead. Along with improved training times, our learned labels are semantically meaningful and can reveal hierarchical relationships that may be present in the data.

\end{abstract}

\section{Introduction}
\label{sec:Introduction}

\begin{figure}
	\centering
	\begin{subfigure}[b]{0.22\textwidth}
		\centering
		\includegraphics[width=\textwidth, height=11em]{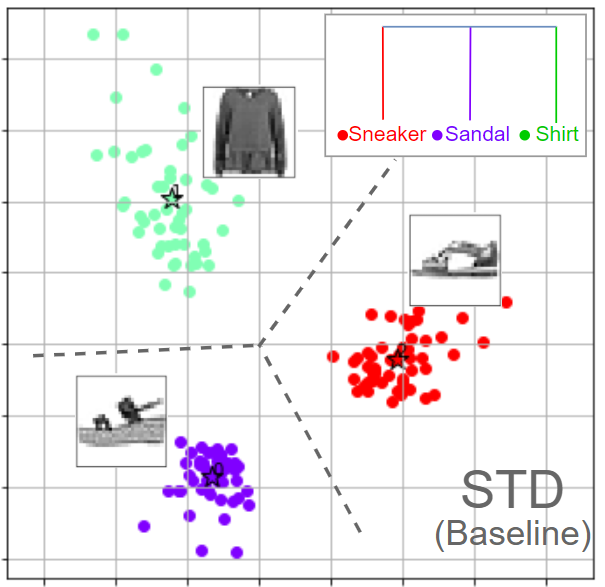}
		\caption{One-hot Labels}
	\end{subfigure}
	\begin{subfigure}[b]{0.22\textwidth}
		\centering
		\includegraphics[width=\textwidth, height=11em]{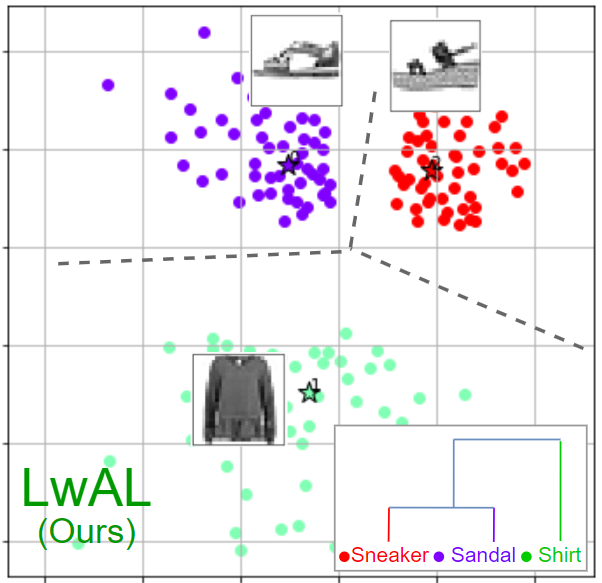}
		\caption{LwAL Learned Labels}
	\end{subfigure}
	\caption{
	A visualization of the labels and how a neural network may map the training examples for a 3-way classification task. Left: when using the one-hot label encoding; Right: when adaptively learning the label encoding.   
    }
	\label{fig:nn_decision_boundary}
\end{figure}

Neural Networks have become an essential tool for achieving high-quality classification in various application domains. Part of their appeal stems from the fact that a practitioner does not have to hand-design the input features for model training. Instead, they can simply use the raw data representation (such as using pixels instead of highly processed SIFT or HOG features for a computer vision task) and learn a mapping to the target class. The high degree of flexibility enables neural networks to learn highly non-linear maps, and thus the target output representation is also usually kept relatively simple. It is customary to encode the target labels as a one-hot encoding\footnote{For a $k$-way classification task, one-hot encoding of the $i$th category is simply the $e_i$ basis vector in $k$ dimensions.}. While simple and computationally convenient, a one-hot representation is rather arbitrary. Indeed, such an encoding destroys any semantic relationships that the target categories may have. For instance, for a 3-class apparel classification task with categories, say, \texttt{sandal}, \texttt{sneaker} and \texttt{shirt}, the semantic similarity between \texttt{sandal} and \texttt{sneaker} (both being \emph{footwear}) is clearly not captured by the one-hot encoding. An alternate label representation can allow us to capture this semantic connection, and perhaps even make the learning process easier (cf.\ Figure \ref{fig:nn_decision_boundary}).

What might be a better representation of the output labels? Since powerful word embedding models such as Word2Vec \citep{word2vec} and BERT \cite{BERT} are known to capture the semantic meaning of commonly occurring words, one can use such prelearned representations of our labels for classification. In fact, \citet{chen2021beyond} explored this idea in detail. They  study the effectiveness of several embeddings including BERT (pretrained on textual data) and audio spectrogram (trained on the vocal pronunciations of the class labels) to represent the target labels and show improved performance.   

An alternate approach, of course, is to explicitly \emph{learn} the label representation from data itself.  
This again can be done in several ways. 
\citet{label-embed-network}, for example, propose to augment the underlying neural network with  specialized layers for data classification and label embedding that interact with each other during the training process. This of course adds complexity to the network potentially increasing network size.
\citet{LWR}, in contrast, learn a ``soft" set of labels by ``smoothing" the original one-hot encodings without modifying the underlying network architecture. While the learned labels are reasonably flexible in representation, a simple smoothing can miss capturing more complex semantic relationships among labels.

In this work, we learn a robust data-dependent label representation that addresses issues that were unresolved in previous literature. We propose Learning with Adaptive Labels (\algo{LwAL}) algorithm, which simultaneously learns semantically appropriate label representations while training for the underlying classification task. \algo{LwAL} is based on the insight that relationships between class labels should be inherent in data belonging to the classes. Since one can view a neural network as a function that maps the input data to a latent representation $\mathbf{z}$, we can utilize this latent data representation to get an initial estimate the label representation $\hat{\mathbf{y}}_{c_i}$ for each class $c_i$. Given the initial estimate of the target labels, we can now tune the underlying neural network parameters to improve classification accuracy. This improved network can in turn, be used to get a better data-dependent label representation in the latent space. We can thus alternate between learning the best representation of the labels in the latent space and learning the best parameters for the underlying network for classification, such that at convergence we achieve both high quality accuracy and an improved representation of the target labels.

\subsection{Our Contributions}
\label{subsec:Contribution}
We propose a simple yet powerful \emph{alternating updates} training algorithm \algo{LwAL}, that can learn high-quality representations of labels. Our algorithm works with any underlying network architecture, without any architecture modifications, and introducing only minimal additional parameters to tune.
We show that learning the labels simultaneously with \algo{LwAL} significantly cuts down on the overall training time (usually by more than 50\% and sometimes up to 80\%) while often achieving better test accuracies than the previous works on label learning. We further show that our learned labels are in fact semantically meaningful and can reveal hierarchical relationships that may be present in our data.

\section{Related Work}
\label{sec:Related Work}
Label representations beyond the one-hot encoding have gained interest in recent years. Here we discuss the related literature in detail.

\subsection{Learning Labels Directly}

\subsubsection{Representations by label smoothing:}\label{subsec:Label Smoothing} Label smoothing techniques aim to modify the hard \emph{one-hot} class probability distribution to a softer target, which can be used to provide broader signals to the model and hence potentially achieving better performance.
Numerous smoothing-based regularization techniques such as Max-Entropy Regularizer \algo{MaxEntReg} \cite{max-entropy}, Teacher-Free Regularizer \algo{TFReg} \cite{TF-reg}, and Learning with Retrospection \algo{LWR} \cite{LWR} have been proposed in the literature, all showing promising improvements. Yet they do not consider unravelling or understanding the relationships between the learned class labels. \citet{LWR} for instance focuses on learning labels generated by a temperature controlled softmax function for better training. Such representations, by their construction, are limited to learning smooth unimodal class probability distributions and cannot capture complex multimodal class distributions that may be necessary to model semantic relationships that may be present in data. 
\subsubsection{Representations by network augmentation:}\label{subsec:network_mod}
\citet{label-embed-network} go beyond just label smoothing and propose a unique approach to augment the underlying neural network with specialized layers to learn sophisticated label representations during the training process.  Interestingly, they show that even though their augmented network is more complex, it usually learns a good classifier at a faster rate, achieving state-of-the-art accuracies for label learning.

\subsubsection{Static Label Representations:}
\label{subsec:static}
Rather than learning a label representation that is tuned to a given classification task, \citet{chen2021beyond} take an alternate approach and use high-quality pre-trained embeddings (such as BERT or GLoVe) to represent their target labels. 
Since no label-training is involved, this approach has the advantage of using good label representations with no added complexity, but suffers from yielding relatively lower classification accuracies.  
This technique also relies on the practitioner having knowledge about which pre-trained embedding is most suitable for the given classification task, which may not be as obvious.

\subsection{Other Notable Related Techniques}
\label{subsec:Other Label Representation Techniques}

While not aiming to learn label representations explicitly, certain ML models yield labels beyond the traditional one-hot encoding as a side-effect.
Student-Teacher learning paradigm \cite{distilling-knowledge}, for instance, aims to learn a more compact network that approximates the behavior of a given large network. In this process of \emph{distillation}, the original one-hot target labels of the larger network usually get an alternate ``dense" representation in the learned compact network. While interesting, learning the distilled network is time-intensive and thus not an efficient mechanism to learn label representations. 

\citet{DEC} develop an unsupervised framework for learning to cluster data in the latent space. They use an auto-encoder architecture to learn a compact latent of the input data where it is forced to form clusters. 
One can thus use these learned latent data clusters and use the cluster centers as a proxy for representing labels. The lack of direct supervision yields suboptimal partitions and hence suboptimal label encodings for classification.

\subsection{Connection to Metric Learning}
\label{subsec:Metric Learning}

Metric learning aims to learn a transformation of the input space where data from the same category is mapped closer together than data from different categories 
\cite{metriclearning_survey_kulis, metriclearning_survey_bellet}.
One can perhaps view learning labels as performing metric learning not on the input space, but rather on the output space. 
Interestingly, to the best of our knowledge, this viewpoint is not explored in existing literature and may be a fruitful avenue for future research. 

Some metric learning literature does explore semantic hierarchical relationships between labels to learn more informed transformations. Notably, \citet{verma-hierarchical} explicitly incorporate label hierarchy information to  
markedly improve nearest-neighbor classification accuracy. They additionally show that such a learned metric can also help in augmenting large taxonomies with new categories. Our work, in contrast, derives the label taxonomy directly from data without any prior hierarchical information.

\section{Methodology} 
\label{sec:Methodology}

Here we formally introduce our Learning with Adaptive Labels \algo{LwAL} algorithm, which simultaneously learns label representations while training for the underlying classification task.
We'll start by reviewing the standard training procedure for neural networks, introducing our notation. We then present our \algo{LwAL} modifications that simultaneously learns the label encodings. Finally we discuss additional optional variations to \algo{LwAL} that can further improve performance in certain applications.

\subsection{Standard Neural Network Training Procedure}
\label{subsec:Reconsidering Standard Training Procedure}
Recall that given a dataset $\mathcal{D}=(X, Y)=\{(x^{(i)}, y^{(i)})\}_{i=1}^{m}$ of $m$ samples for a $N$-category classification task, where $x^{{(i)}}$ denotes the application specific input representation and $y^{(i)}$ denotes the one-hot output representation of the $i$-th sample, the goal of a neural network $f_\theta$ (parameterized by $\theta$) to learn a mapping from the inputs ($x^{{(i)}}$) to the outputs ($y^{(i)}$). 
This learning is usually done by finding a parameter setting $\theta$ that minimizes the loss between the predicted output $f_\theta(x^{(i)})$ and the desired (one-hot) output $y^{(i)}$. In particular, let $z^{(i)} = f_\theta(x^{(i)})$ be the network encoding of the input $x^{(i)}$. First a Softmax is applied to $z^{(i)}$ to obtain a probability distribution which encodes the affinity of $z^{(i)}$ to each of the $N$ classes. Then this induced probability distribution is compared with the ideal probability distribution $y^{(i)}$ using any distribution-comparing divergence such as the cross-entropy (CE). Thus the classification loss for the $i$-th sample becomes  
\begin{align*}
\mathcal{L}_{\mathrm{cls} }^{(i)}(\theta) :=& \;\; \mathrm{CE}(y^{(i)}, \mathrm{Softmax}(z^{(i)})) \\ =& \;\;
\mathrm{CE}(y^{(i)},\mathrm{Softmax}(f_\theta(x^{(i)}) ). 
\end{align*}
The optimal parameter setting $\theta$ can thus be learned by usual iterative gradient-type updates (such as SGD or Adam) on the aggregate loss over all training datapoints.

\subsection{Learning with Adaptive Labels}
\label{subsec:Learning with Adaptive Labels}

To learn more enriched, semantically meaning label representations, we posit that that semantic relationships between classes are contained within the samples belonging to the class. Specifically, we model the label representation $\hat{\mathbf{y}}_{c_j}$ of a class $c_j$ as the vector that minimizes the average distance to the network encoding of the samples $z^{(i)}$ belonging to class $c_j$. This is equivalent to considering  
\begin{equation}\label{eq:centroid_eq}
	\mathbf{\hat{y}}_{c_j} := \frac{1}{m_{c_{j}}} \sum\limits_{y^{(i)}=c_{j}} \mathbf{z}^{(i)},
\end{equation}
where $m_{c_j}$ is the number of samples belonging to class $c_j$. 

To bring the training in line with standard neural network updates, given this new class representation, one can define the probability that the network encoding $z^{(i)}$ of the $i$-th datapoint belonging to class $c_j$ as 
\begin{equation}\label{eq:LwAL_prediction}
	p^{(i)}_j  
	:= \mathrm{Softmax}\left(-\left\| \mathbf{z}^{(i)}-\mathbf{\hat{y}}_{c_j} \right\|_2 \right).
\end{equation}
Therefore, the modified cross entropy loss for the $i$-th datapoint becomes
\begin{align*}\label{eq:lwal_loss}
\mathcal{L}_{\mathrm{LwAL} }^{(i)}(\theta) :=& \;\; \mathrm{CE}(y^{(i)}, \mathbf{p}^{(i)}),
\end{align*}
where $\mathbf{p}^{(i)} = (p^{(i)}_j)_j$ is the probability distribution that encodes the affinity of $z^{(i)}$ to each of the $k$ classes using the new label representation.
One can thus train the optimal parameters of the underlying neural network $f_\theta$ the usual way.

One should note that the choice of cross-entropy as the loss function encourages the learned class representations $\mathbf{\hat{y}}_{c_i}$ to be well separated yielding empirically better accuracies than other popular loss functions. 

One can predict the label of test examples $x_\textup{test}$ by simply assigning it to the closest learned label in the network encoded space. That is 
	\[
		\hat{y}_\textup{test} = \arg\min_{c_i} \left\| f_\theta(x_\textup{test}) - \mathbf{\hat{y}}_{c_i}  \right\|_2.
	\]

\subsubsection{Adapting to large-scale datasets} To accommodate large scale datasets, we use the mini-batch paradigm. The mini-batch training usually suffers from the problem of \emph{moving target} \citep{DQN}, that is, $\mathbf{\hat{y}}_{c_i}$ are constantly changing leading to poor convergence. In order to alleviate this, we add hyperparameters $k$ that controls the update frequency \citep{LWR}, and initial warmup steps $w$ to promote more initial separation between classes when learning $\mathbf{\hat{y}}_{c_i}$.
See Algorithm \ref{alg:LwAL_training_algo} for details.

\begin{algorithm}
	\caption{LwAL Training Algorithm}\label{alg:LwAL_training_algo}
	\begin{algorithmic}[1]
	\Require input dataset $(\mathbf{X}, \mathbf{Y}) \sim D$
	\Require neural network $f_\theta$
	\Require number of training steps per epoch $n$
	\Require update frequency $k\ge 1$
	\Require warmup steps $w \ge 0$
	\Repeat 
	\For{step $i=1, \ldots , n$}
		\State sample a large batch $(\mathbf{x}, \mathbf{y}) \sim D$
		\State $\mathbf{z}\gets f_\theta(\mathbf{x})$
		\State compute $\mathcal{L}= \mathcal{L}_{\textup{LwAL}}$
		\If{$i > w$ and $(i-w)\mod k = 1$}
			\State update $\mathbf{\hat{y}}_{c_i}$ for each class $c_{i}$ as per Eq.\ (\ref{eq:centroid_eq})
			\State compute $\mathcal{L}= \mathcal{L}_{\textup{LwAL}}\;\; \underbrace{+\lambda \cdot \mathcal{L}_{\mathrm{repel}}}_{\textup{optional}}$
		\EndIf
		\State gradient descent on $\mathcal{L}$ to update $\theta$
	\EndFor
	\Until{convergence}
	\end{algorithmic}
\end{algorithm}

\subsection{Additional Improvements} To further improve the label quality, we draw inspiration from the push-pull based losses from metric learning literature \cite{MMC, LMNN, FaceNet}. We add an optional ``push" loss, that encourages our learned labels to be well-separated thus yielding better generalization accuracies. 
Specifically, we penalize the angle between the network encoding of the datapoints $z^{(i)}$ from different classes, using cosine similarity. That is (c.f.\ Algorithm \ref{alg:LwAL_training_algo}),
\begin{equation}\label{eq:repel_loss}
	\mathcal{L}_{\mathrm{repel}}(\theta) := \sum\limits_{i,j}  \mathbbm{1}\{ y^{(i)} \neq y^{(j)} \}\cdot \mathrm{sim}_{\cos}(z^{(i)},z^{(j)}).
\end{equation}

\section{Experiments}
\label{sec:Experiments}

We have a two-fold aim for our empirical study.  First, we evaluate how \algo{LwAL} fares (both in terms of speed and accuracy) when compared to other popular label-learning methodologies on benchmark datasets. Second, we evaluate the effectiveness of our learned labels for revealing semantically meaningful categorical relationships in our data.\footnote{An implementation of our algorithm is available at \texttt{https://github.com/jasonyux/Learning-with-} \texttt{Adaptive-Labels}.}

\begin{table*}[!th]
	\centering
	\begin{tabular}{|cc|ccc|ccc|}
		\hline
		\multicolumn{2}{|c|}{}
		&\multicolumn{3}{|c|}{Percent Time/Epoch Reduced}
		&\multicolumn{3}{|c|}{Best Test Accuracy}\\
&&ResNet50  &EfficienNetB0 &DenseNet121  &ResNet50  &EfficienNetB0 &DenseNet121 \\
		\hline
\multirow{9}{*}{
\rotatebox[origin=c]{90}{MNIST}
}
&One-hot (STD)     
& \multicolumn{3}{|c|}{Reference} &99.1$\pm$0.1 &99.4$\pm$0.0 &99.1$\pm$0.1 \\

&StaticLabel  
&-    &- &-    &N/A    &N/A   &N/A   \\

&LWR2k  
& - &\textbf{70\%} &50\% &99.2$\pm$0.0 &99.4$\pm$0.1 &99.3$\pm$0.1 \\

&LWR3k  
&50\% &50\% &40\% &99.1$\pm$0.1 &\textbf{99.5$\pm$0.1} &99.2$\pm$0.1 \\

&LWR5k  
&10\% &30\% &30\% &99.2$\pm$0.1 &99.4$\pm$0.1 &99.2$\pm$0.1\\

&LabelEmbed  
&50\% &10\% &60\% &99.2$\pm$0.1 &99.4$\pm$0.0 &\textbf{99.4$\pm$0.0}\\

&\cellcolor{gray!25}LwAL (Ours) 
&\cellcolor{gray!25} \textbf{60\%} & \cellcolor{gray!25}20\% &\cellcolor{gray!25} - &\cellcolor{gray!25}99.2$\pm$0.1 &\cellcolor{gray!25}99.3$\pm$0.1 &\cellcolor{gray!25}99.2$\pm$0.0\\

&\cellcolor{gray!25} LwAL10 (Ours) 
&\cellcolor{gray!25} \textbf{60\%} &\cellcolor{gray!25} - &\cellcolor{gray!25} - &\cellcolor{gray!25}\textbf{99.3$\pm$0.1} &\cellcolor{gray!25}99.3$\pm$0.0 &\cellcolor{gray!25}99.1$\pm$0.0\\

&\cellcolor{gray!25}LwAL10+rpl (Ours) 
&\cellcolor{gray!25}50\% &\cellcolor{gray!25} - &\cellcolor{gray!25}\textbf{70\%} &\cellcolor{gray!25}\textbf{99.3$\pm$0.1} &\cellcolor{gray!25}99.3$\pm$0.0 &\cellcolor{gray!25}\textbf{99.4$\pm$0.0}\\
	\hline\hline
\multirow{9}{*}{ \rotatebox[origin=c]{90}{Fashion MNIST}} 
&One-hot (STD)      
&\multicolumn{3}{|c|}{Reference} &92.3$\pm$0.2 &93.1$\pm$0.2 &92.4$\pm$0.3 \\

&StaticLabel  
&30\% & - &20\% &92.8$\pm$0.1 &93.0$\pm$0.1 &92.6$\pm$0.2\\

&LWR2k  
& - &20\% & - &92.3$\pm$0.2 &93.1$\pm$0.2 &92.4$\pm$0.3 \\

&LWR3k  
& - &\textbf{50\%} &20\% &92.1$\pm$0.0 &93.3$\pm$0.3 &92.2$\pm$0.4\\

&LWR5k  
& - &40\% &30\% &92.1$\pm$0.0 &\textbf{93.4$\pm$0.1} &92.3$\pm$0.4\\

&LabelEmbed  
&40\% &10\% &\textbf{60\%} &92.7$\pm$0.4 &93.1$\pm$0.2 &92.9$\pm$0.1\\

&\cellcolor{gray!25}LwAL (Ours) 
&\cellcolor{gray!25}\textbf{50\%} &\cellcolor{gray!25} - &\cellcolor{gray!25}30\% &\cellcolor{gray!25}\textbf{92.9$\pm$0.1} &\cellcolor{gray!25}93.0$\pm$0.2 &\cellcolor{gray!25}92.4$\pm$0.0\\

&\cellcolor{gray!25}LwAL10 (Ours) 
&\cellcolor{gray!25}\textbf{50\%} &\cellcolor{gray!25} - &\cellcolor{gray!25}40\% &\cellcolor{gray!25}92.3$\pm$0.0 &\cellcolor{gray!25}92.7$\pm$0.2 &\cellcolor{gray!25}92.6$\pm$0.2\\

&\cellcolor{gray!25}LwAL10+rpl (Ours) 
&\cellcolor{gray!25}30\% &\cellcolor{gray!25} - &\cellcolor{gray!25}\textbf{60\%} &\cellcolor{gray!25}92.7$\pm$0.2 &\cellcolor{gray!25}92.8$\pm$0.2 &\cellcolor{gray!25}\textbf{93.0$\pm$0.2}\\
	\hline\hline
\multirow{9}{*}{
\rotatebox[origin=c]{90}{CIFAR10}
} 
&One-hot (STD)      
& \multicolumn{3}{|c|}{Reference} &73.3$\pm$0.5 &75.9$\pm$0.4 &78.8$\pm$0.5 \\

&StaticLabel  
& - & - & - &74.0$\pm$0.7 &75.7$\pm$0.5 &77.7$\pm$0.3\\

&LWR2k  
& - & - & - &67.8$\pm$1.1 &74.7$\pm$0.3 &74.1$\pm$0.7 \\

&LWR3k  
& - & - & - &69.3$\pm$0.8 &75.3$\pm$0.1 &75.6$\pm$0.9\\

&LWR5k  
& - &30\% & - &69.9$\pm$1.1 &76.3$\pm$0.4 &76.9$\pm$0.7\\

&LabelEmbed  
& - &30\% &40\% &72.2$\pm$0.9 &76.7$\pm$0.3 &79.4$\pm$0.4\\

&\cellcolor{gray!25}LwAL (Ours) 
&\cellcolor{gray!25} - &\cellcolor{gray!25}20\% & \cellcolor{gray!25}- &\cellcolor{gray!25}72.8$\pm$0.2 &\cellcolor{gray!25}76.7$\pm$0.4 &\cellcolor{gray!25}78.9$\pm$0.0 \\

&\cellcolor{gray!25}LwAL10 (Ours) 
&\cellcolor{gray!25}30\% &\cellcolor{gray!25} - &\cellcolor{gray!25}30\% &\cellcolor{gray!25}74.3$\pm$0.3 &\cellcolor{gray!25}76.2$\pm$0.2 &\cellcolor{gray!25}79.2$\pm$0.4\\

&\cellcolor{gray!25}LwAL10+rpl (Ours) 
&\cellcolor{gray!25}\textbf{60\%} &\cellcolor{gray!25}\textbf{50\%} &\cellcolor{gray!25}\textbf{50\%} &\cellcolor{gray!25}\textbf{76.0$\pm$0.4} &\cellcolor{gray!25}\textbf{77.9$\pm$0.5} &\cellcolor{gray!25}\textbf{80.5$\pm$0.3}\\
	\hline\hline
\multirow{9}{*}{
\rotatebox[origin=c]{90}{CIFAR100}
}
&One-hot (STD)      
& \multicolumn{3}{|c|}{Reference} &37.4$\pm$0.6 &40.5$\pm$0.5 &44.6$\pm$0.8\\

&StaticLabel  
& - & - & - &16.8$\pm$1.1$^{*}$ &5.9$\pm$0.6 &7.8$\pm$0.3$^{*}$\\

&LWR2k  
& - & - & - &32.9$\pm$0.5 &38.1$\pm$0.5 &38.7$\pm$0.4\\

&LWR3k  
& - & - & - &32.7$\pm$0.1 &38.1$\pm$0.4 &38.6$\pm$0.6\\

&LWR5k  
& - & - & - &32.9$\pm$0.4 &38.1$\pm$0.7 &38.6$\pm$0.6\\

&LabelEmbed  
&10\% &20\% &- &37.7$\pm$0.8 &41.0$\pm$0.5 &44.6$\pm$0.7\\

&\cellcolor{gray!25}LwAL (Ours) 
&\cellcolor{gray!25}\textbf{70\%} &\cellcolor{gray!25}\textbf{65\%} &\cellcolor{gray!25}60\% &\cellcolor{gray!25}38.8$\pm$0.4 &\cellcolor{gray!25}\textbf{43.2$\pm$0.2} &\cellcolor{gray!25}46.8$\pm$0.3\\

&\cellcolor{gray!25}LwAL10 (Ours) 
&\cellcolor{gray!25}\textbf{70\%} &\cellcolor{gray!25}50\% &\cellcolor{gray!25}\textbf{65\%} &\cellcolor{gray!25}39.3$\pm$0.2 &\cellcolor{gray!25}41.6$\pm$0.6 &\cellcolor{gray!25}47.5$\pm$0.4\\

&\cellcolor{gray!25}LwAL10+rpl (Ours) 
&\cellcolor{gray!25}\textbf{70\%} &\cellcolor{gray!25}60\% &\cellcolor{gray!25}60\% &\cellcolor{gray!25}\textbf{39.9$\pm$0.4} &\cellcolor{gray!25}42.2$\pm$0.5 &\cellcolor{gray!25}\textbf{48.0$\pm$0.0}\\
	\hline\hline
\multirow{9}{*}{
\rotatebox[origin=c]{90}{FOOD101}
}
&One-hot (STD)      
& \multicolumn{3}{|c|}{Reference} &16.3$\pm$0.3 &18.5$\pm$0.5 &20.6$\pm$0.0\\

&StaticLabel  
& - & - & - &2.6$\pm$0.5$^{*}$ &2.0$\pm$0.8 &6.4$\pm$0.5$^{*}$\\

&LWR2k  
& - & - & - &13.8$\pm$0.1 &18.0$\pm$0.3 &17.9$\pm$0.2\\

&LWR3k  
& - & - & - &13.9$\pm$0.1 &18.2$\pm$0.3 &18.0$\pm$0.3\\

&LWR5k  
& - & - & - &13.9$\pm$0.1 &18.5$\pm$0.4 &17.8$\pm$0.1\\

&LabelEmbed  
&10\% &65\% & 35\% &15.8$\pm$0.1 &19.8$\pm$0.3 &21.6$\pm$0.5 \\

&\cellcolor{gray!25}LwAL (Ours) 
&\cellcolor{gray!25}75\% &\cellcolor{gray!25}\textbf{80\%} &\cellcolor{gray!25}70\% &\cellcolor{gray!25}16.6$\pm$0.3 &\cellcolor{gray!25}\textbf{22.0$\pm$0.6} &\cellcolor{gray!25}21.1$\pm$0.1 \\

&\cellcolor{gray!25}LwAL10 (Ours) 
&\cellcolor{gray!25}\textbf{80\%} &\cellcolor{gray!25}\textbf{80\%} &\cellcolor{gray!25}\textbf{75\%} &\cellcolor{gray!25}17.5$\pm$0.3 &\cellcolor{gray!25}20.5$\pm$0.5 &\cellcolor{gray!25}22.5$\pm$0.1\\

&\cellcolor{gray!25}LwAL10+rpl (Ours) 
&\cellcolor{gray!25}\textbf{80\%} &\cellcolor{gray!25}\textbf{80\%} &\cellcolor{gray!25}70\% &\cellcolor{gray!25}\textbf{17.7$\pm$0.1} &\cellcolor{gray!25}20.9$\pm$0.2 &\cellcolor{gray!25}\textbf{22.9$\pm$0.1}\\
	\hline
\end{tabular}
\caption{Learning accuracy and speed comparison between \algo{LwAL} and other baselines. 
\algo{LwAL} is trained using 0 warmup steps and update frequency of once per step. Blank (--) indicates cases when the specific algorithm+backbone pair was unable to reach the reference \algo{STD} test accuracy. Star (*) indicates the use of different learning rate ( $1e^{-3}$) due to failure of convergence. N/A for \dataset{MNIST} dataset using \algo{StaticLabel} indicates that the BERT representation of \dataset{MNIST} categories is not appropriate.}
\label{tbl:image_dset_resnet_performance_comparison}
\end{table*}

\begin{figure*}[!htb]
	\centering
	\begin{subfigure}[b]{0.85\textwidth}
		\centering
		\includegraphics[width=\textwidth, height=24em]{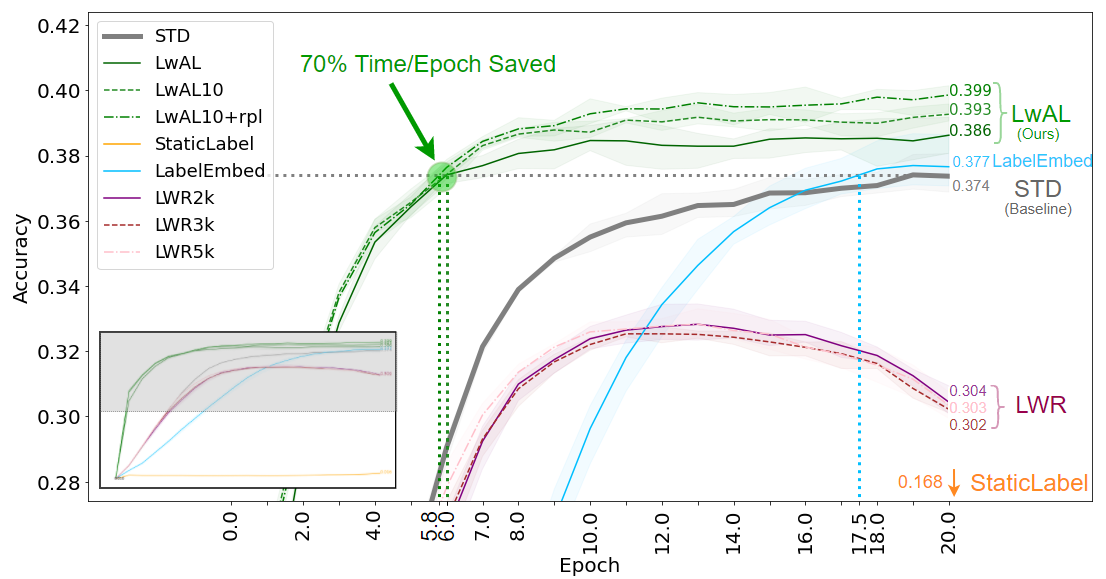}
		\caption{Using ResNet50 backbone}
	\end{subfigure}
	\begin{subfigure}[b]{0.4\textwidth}
		\centering
		\includegraphics[width=\textwidth, height=12em]{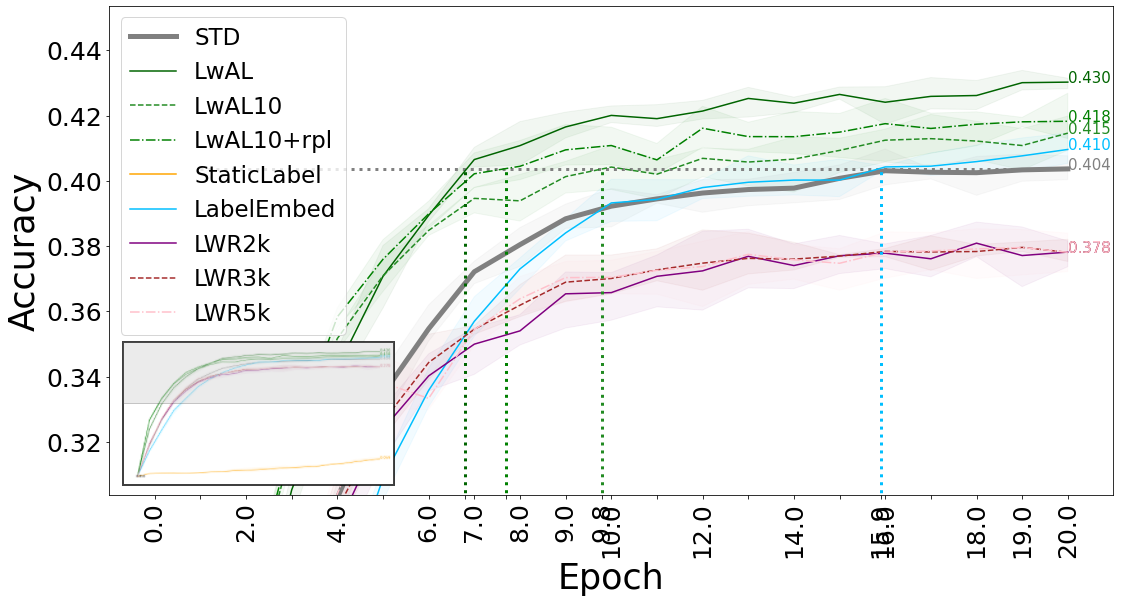}
		\caption{Using EfficientNetB0 backbone}
	\end{subfigure}
	\begin{subfigure}[b]{0.4\textwidth}
		\centering
		\includegraphics[width=\textwidth, height=12em]{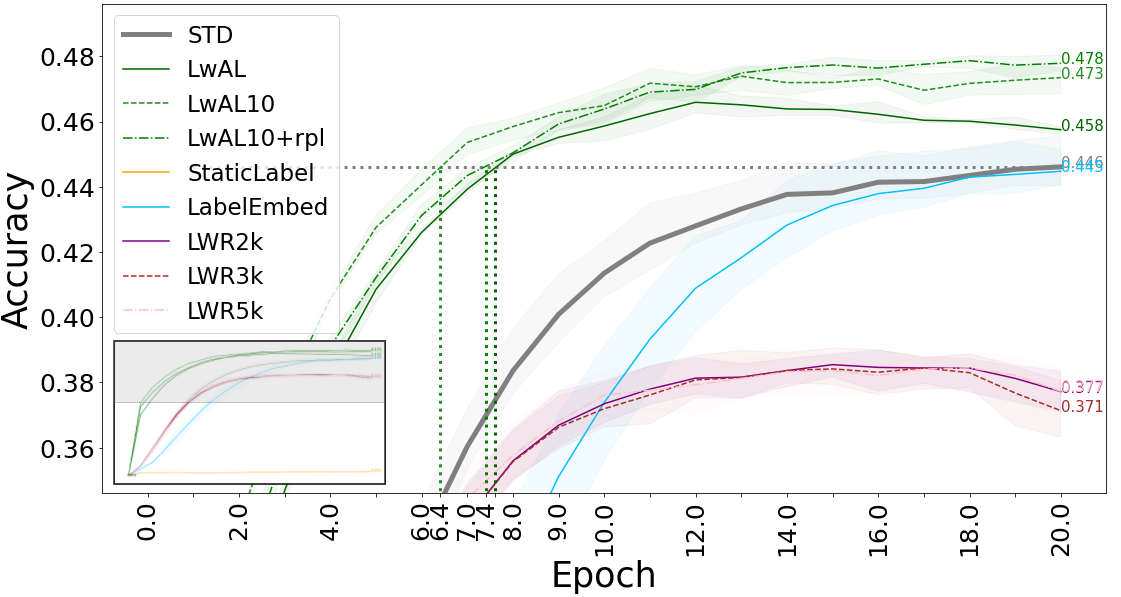}
		\caption{Using DenseNet121 backbone}
	\end{subfigure}
	\caption{Test Curves for LwAL and other baseline algorithms trained on CIFAR100 dataset. STD best accuracy is used as a reference for other algorithms.}
	\label{fig:LwAL_test_curves}
\end{figure*}

\subsection{Learning Speed and Test Performance}
\label{subsec:Learning Speed and Test Performance}

\subsubsection{Datasets}
\label{subsubsec:Datasets}
To evaluate the robustness of our technique, we report results on several benchmark datasets with different sizes, number of categories and application domains. In particular we used the following datasets for our experiments.
\begin{center}
\begin{tabular}{ |c|c|c|c| } 
 \hline
 Dataset & Domain & \# classes & \# points \\ 
 \hline
 \dataset{MNIST} & Vision & 10 & 60k \\ 
 \dataset{Fashion MNIST} & Vision & 10 & 60k \\ 
 \dataset{CIFAR10} & Vision & 10 & 50k \\ 
 \dataset{CIFAR100} & Vision & 100 & 50k \\ 
 \dataset{FOOD101} & Vision & 101 & 25k \\ 
 \dataset{IMDB Reviews} & NLP & 2 & 25k \\ 
 \dataset{YELP Reviews} & NLP & 2 & 560k \\ 
 \hline
\end{tabular}
\end{center}
We use the default train/test splits provided by the tensorflow library as of Aug 2022.

\nocite{mnist, fmnist, cifar, food101, imdb, yelp}

\subsubsection{Network Architectures}
\label{subsubsec:Architectures}
To check if \algo{LwAL} works across different architectures, we test on ResNet50 \cite{resnet50}, EfficientNetB0 \cite{efficientnet}, and DenseNet121 \cite{densenet} with ImageNet weights, for vision datasets. All of these architectures are available on the tensorflow library as of Aug 2022. For text datasets, we use BERT \cite{BERT} which is available on the huggingface library as of Aug 2022.

\subsubsection{Baselines}
\label{subsubsec:Baselines}
We compare \algo{LwAL} with several important baselines. We compare with the standard one-hot training procedure (\algo{STD}). \citet{chen2021beyond} employ a static pretrained (BERT or audio spectrogram) label representation (\algo{StaticLabel}). For our comparisons, we chose the pretrained BERT embedding as it was reported to show good performance on the benchmark datasets. From the label smoothing techniques, we use \algo{LWR} \cite{LWR} with varying choices of the update-frequency hyperparameter ($k=2, 3,5$). We also compare with the network augmentation (\algo{LabelEmbed}) technique by \citet{label-embed-network}.

\subsubsection{Hyperparameters}
\label{subsubsec:Hyperparameters}
In order for the backbones (ResNet50, EfficienNetB0, DenseNet121) to be used across different datasets, we attach a single dense layer with $l_2$ regularization of $0.1$ at the top to be used as the classification head.

We train all algorithms with the same set of parameters for consistency. We first pick a learning rate within the same backbone so that all algorithms can converge: 
for ResNet50 and DenseNet122, we use ADAM optimizer with $\beta_1=0.9, \beta_2=0.999$ and learning rate of $0.0001$; for EfficientNetB0, we use the same optimizer but with learning rate of $0.001$. For small datasets such as \dataset{MNIST}, \dataset{F.MNIST}, and \dataset{CIFAR10}, we train all algorithms over 10 epochs. For large datasets such as \dataset{CIFAR100} and \dataset{FOOD101}, we train all algorithms over 20 epochs where we see the test accuracy reaches a plateau and starts to overfit. We repeat all runs with seeds $12,123,1234$ and report the mean and spread.

For \algo{LWR}, we use temperature $\tau=5$, which is the recommended value. Since we are only training for a few epochs, we also experiment with varying values for the frequency $k=2,3,5$ and report all results in Table \ref{tbl:image_dset_resnet_performance_comparison}. For \algo{LabelEmbed}, we use the default setting of the parameters in the implementation    \citep{label-embed-network} (i.e.\ $\tau=2$, $\alpha=0.9$, and $\beta=0.5$). 

For \algo{LwAL}, we can vary the output label dimension. We compare the results for output dimension of 10 times\footnote{We empirically found that increasing the output dimension often leads to improved performance, as discussed by \citet{chen-etal-2020-label}. Empirically, 10 times the number of classes usually leads to best performance.} the number of classes (\algo{LwAL10}). We also compare the results with the addition of optional loss $\mathcal{L}_{\mathrm{repel}}$ (\algo{LwAL10+rpl}). We use update frequency of $k=1$ and no warmup steps as we use large batch sizes. For \algo{LwAL10+rpl} we used $\lambda=10$ (cf.\ Algorithm \ref{alg:LwAL_training_algo}).

\subsubsection{Results and Observations}
\label{subsubsec:Learning Speed Results}

Tables \ref{tbl:image_dset_resnet_performance_comparison} (main text) and \ref{tbl:text_dset_comparison} (Appendix) summarize our results for the Vision and NLP datasets respectively. Best results are highlighted in bold. Blank (--) in the Time column indicates that a particular algorithm+backbone combination was not able to achieve the \algo{STD} one-hot baseline test accuracy.  

Observe that \algo{LwAL} significantly cuts down on the overall training time (usually by more than 50\% and sometimes up to 80\%) while often achieving better test accuracies over other baselines. Figure \ref{fig:LwAL_test_curves} depicts how the test accuracy curve improves as the training proceeds for a typical run using various backbones. It clearly highlights that one can achieve the same test accuracy as \algo{STD} with 70\% reduction in training time. This phenomenon is typical for various benchmark datasets and choice of backbones (cf.\ Table \ref{tbl:image_dset_resnet_performance_comparison}). One can conclude that \algo{LwAL10+repl} with DenseNet121 backbone seems to give the best 
results with significant ($\geq 50\%$) savings overall.
Curiously \algo{StaticLabel} and \algo{LWR} are not able to achieve \algo{STD} one-hot label test accuracies for large multi-class datastes like \dataset{CIFAR101} and \dataset{FOOD101}.

\subsection{Semantic Label Representation}
\label{subsec:Semantic Label Representation}
Here we want to empirically evaluate the effectiveness of our learned labels in discovering semantic relationships among categories. For this, we shall use the semantic hierarchy induced by WordNet \cite{WordNet} as the gold standard relationship among the categories, and compare how well our learned labels reveal those relationships. 

To this end, we utilize the Kendall's Tau-b ($\tau_b$) correlation coefficient score to compare the learned representations with the WordNet hierarchy. Specifically, first we compute the pairwise distances between distinct class labels for (i) the reference WordNet hierarchy tree (this is done using the short path distances between the tree nodes) $d_{\mathrm{WN}}(c_i, c_j)$, and (ii) the learned vectors from the label learning algorithm $d_{\mathrm{LwAL}}(c_i, c_j)$. Next, treating collected distance vectors  $\mathbf{d}({c_i}) := (d(c_i, c_j))_{j=1}^N$ (where $i\neq j$) for each of the $N$ classes as rank vectors, we can compute the average semantic correlation score as:
\begin{equation}\label{eq:tau_b_correlation}
\textup{corr(\algo{LwAL})} := \frac{1}{N} \sum\limits_{i=1}^N \tau_b\left(\mathbf{d}_{\mathrm{WN}}({c_i}) , \mathbf{d}_{\mathrm{LwAL}}({c_i}) \right)
\end{equation}

\subsubsection{Datasets}
\label{subsubsec:Semantic Label Datasets}
We report results on datasets for which the classes can be easily mapped to the WordNet \cite{WordNet} hierarchy. This includes the existing \dataset{Fashion MNIST} (8 out of 10 classes can be mapped) and \dataset{CIFAR10} (10 out of 10 classes can be mapped). We also include the results for Animal with Attributes 2 (\dataset{AwA2}) dataset \cite{AwA2}, where 23 out of 50 classes can be mapped). We learn and evaluate the quality of the label representations of only the mappable classes for each of these datasets.

\subsubsection{Architectures, Hyperparameters, and Baselines}
\label{subsubsec:Semantic Label Architectures, Hyperparameters, and Baselines}
We use ResNet50 (with ImageNet weights) as the underlying neural network backbone for our experiments. We compare the results of our \algo{LwAL} algorithm with other label learning techniques: \algo{LWR} (best across $k\in\{2,3,5\}$) and \algo{LabelEmbed}. 
For \algo{LWR}, the explicit label representation is computed via Eq.\ \eqref{eq:centroid_eq}. 
For \algo{LabelEmbed}, since it returns a similarity matrix between the learned labels, we compute the vectorial representation the standard (eigendecomposition) way. 

The rest of the hyperparameter settings (including random seed, batch size, etc.) are same as the previous section.

\begin{table*}[!th]
	\centering
	\begin{tabular}{|c|c|c|c|c|c|}
		\hline
		\multicolumn{1}{|c|}{Datasets}
		&\multicolumn{2}{|c|}{Other Label Learning Algs.}
		&\multicolumn{3}{|c|}{Ours}\\
		\hline
	  &\algo{LWR}&\algo{LabelEmbed}&\algo{LwAL} &\algo{LwAL10} &\algo{LwAL10+rpl}\\
\dataset{CIFAR10}
&-0.017$\pm$0.068 &0.053$\pm$0.058 &0.473$\pm$0.028 &0.544$\pm$0.024 &\textbf{0.609$\pm$0.019}\\

\dataset{F.MNIST} 
&0.019$\pm$0.068 &0.079$\pm$0.172  &0.306$\pm$0.056 &\textbf{0.494$\pm$0.054} &0.305$\pm$0.039 \\

\dataset{AwA2}	
&-0.097$\pm$0.074 &0.088$\pm$0.078  &\textbf{0.299$\pm$0.021} &0.288$\pm$0.024 &0.260$\pm$0.030 \\
		\hline
	\end{tabular}
\caption{Structure correlation score (Eq.\ \ref{eq:tau_b_correlation}) between learned labels and WordNet. 
Bold indicates best performance.}
\label{tbl:LwAL_hierarchy_correlation}
\end{table*}

\begin{figure*}[t]
\centering
		\includegraphics[width=0.36\textwidth]{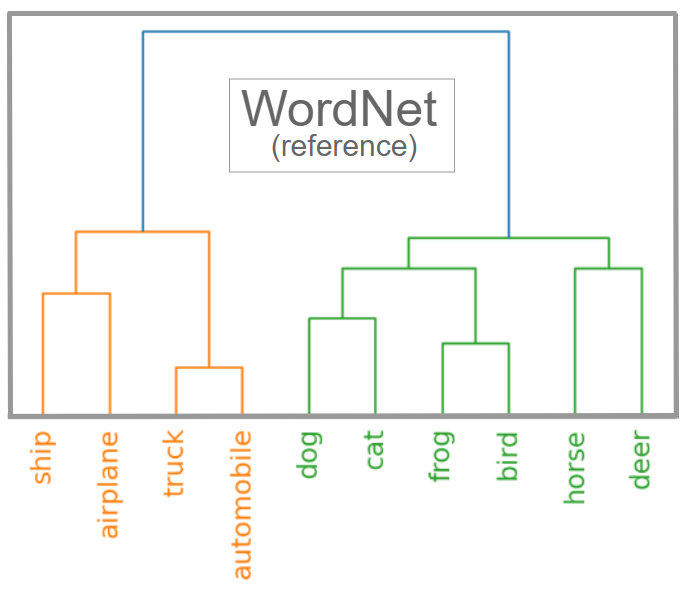}
				\includegraphics[width=0.35\textwidth]{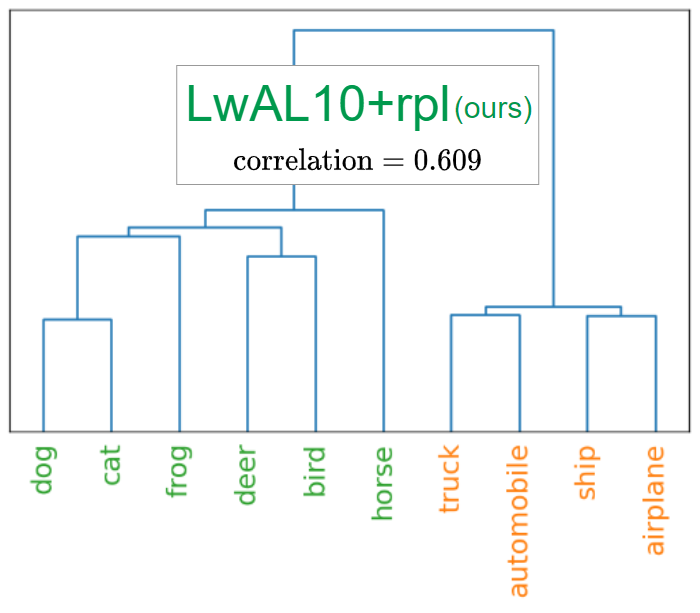} \\
				
			\includegraphics[width=0.35\textwidth]{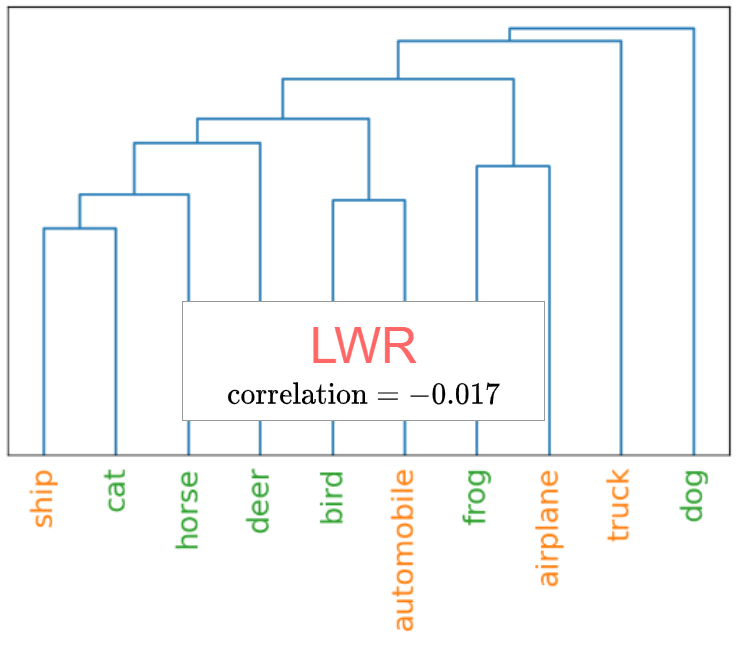}
		\includegraphics[width=0.35\textwidth]{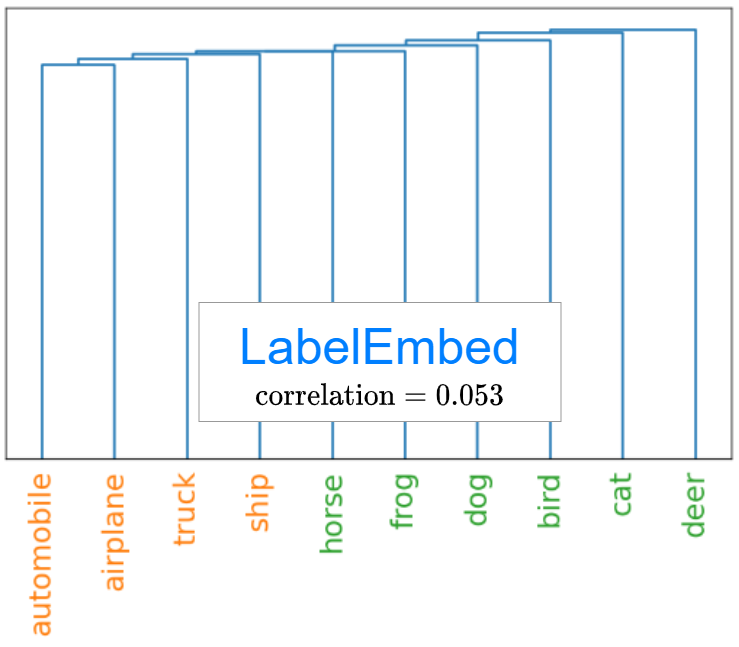}		

	\caption{Hierarchical visualization (via average-linkage) of the learned labels for different algorithms on \dataset{CIFAR10} dataset.}
	\label{fig:LwAL_learnt_structure}
\end{figure*}

\subsubsection{Results}
\label{subsubsec:Semantic Label Results}
We present the correlation score for each of the label learning techniques in Table \ref{tbl:LwAL_hierarchy_correlation}, and and example visualization of the learned hierarchy in Figure \ref{fig:LwAL_learnt_structure}.

Observe that \algo{LwAL} and its variants can consistently generate significantly superior semantically meaningful representations when compared to other label learning methods.
While these results are compelling, it is worth noting that the learned labels and thus the semantic hierarchy is derived from the data inputs. \algo{LwAL} can thus only extract those relationships that are present in the input data representation and likely cannot  capture every fine-grained semantic detail between classes.
Indeed, if the input representation (for example pixels for image classification tasks) does not contain any information about the semantic relationships, then one cannot expect \algo{LwAL}  to capture any useful relationship.

\section{Conclusion and Future Work}
\label{sec:Conclusion and Future Work}
In this work we present a simple yet powerful Learning with Adaptive Labels \algo{LwAL} algorithm that can learn semantically meaningful label representations that the vanilla one-hot encoding is unable to capture.
Interestingly, we find that by allowing the network to flexibly learn a label representation during training, we can significantly cut down on the overall training time while achieving high test accuracies. 
 Extensive experiments on multiple datasets with varying dataset sizes, application domains, and  network architectures show that our learning algorithm is effective and robust.
 
As noted, although \algo{LwAL} can learn high-level semantically meaningful label representations extracted from inputs, it is interesting to explore to what degree this is possible. Can fine-grained semantic relationships be derived just from the raw input space? Or does one need to incorporate additional ``side-information" to accelerate semantic discovery?
We leave this as a topic for future research.

\bibliography{citations}

\newpage
\newpage
\section{Appendix}
\label{sec:Appendix}

\begin{table*}[!th]
	\centering
	\begin{tabular}{|cc|c|c|c|}
		\hline
		\multicolumn{2}{|c|}{}
		&\multicolumn{1}{|c|}{Pecent Time/Epoch Reduced}
		&\multicolumn{1}{|c|}{Best Test Accuracy}
		&\multicolumn{1}{|c|}{Avg. AUAC}\\
		\hline
\multirow{9}{*}{\rotatebox[origin=c]{90}{\dataset{IMDB}}} 
&One-hot (STD)      
&Reference &83.9$\pm$0.1 &82.9$\pm$0.2 \\
&StaticLabel  
&20\% &84.1$\pm$0.1 &82.1$\pm$0.1\\
&LWR2k  
&- &82.8$\pm$0.2 &79.8$\pm$0.4 \\
&LWR3k  
&- &83.2$\pm$0.2 &80.2$\pm$0.2 \\
&LWR5k  
&- &83.7$\pm$0.0 &82.0$\pm$0.2 \\

&LabelEmbed  
&- &82.7$\pm$0.1 &81.5$\pm$0.1 \\

&\cellcolor{gray!25}LwAL (Ours) 
&\cellcolor{gray!25}30\% &\cellcolor{gray!25}\textbf{84.3$\pm$0.1} &\cellcolor{gray!25}83.5$\pm$0.2\\

&\cellcolor{gray!25}LwAL10 (Ours) 
&\cellcolor{gray!25}- &\cellcolor{gray!25}83.8$\pm$0.1 &\cellcolor{gray!25}83.3$\pm$0.1 \\
&\cellcolor{gray!25}LwAL10+rpl (Ours) 
&\cellcolor{gray!25}\textbf{60\%} &\cellcolor{gray!25}84.2$\pm$0.1 &\cellcolor{gray!25}\textbf{83.6$\pm$0.0}\\
	\hline\hline
\multirow{9}{*}{\rotatebox[origin=c]{90}{\dataset{YELP}}} 
&One-hot (STD)      
&- &88.5$\pm$0.1 &88.0$\pm$0.0\\
&StaticLabel  
&60\% &88.6$\pm$0.0 &88.3$\pm$0.0 \\
&LWR2k  
&60\% &\textbf{88.7$\pm$0.2} &\textbf{88.5$\pm$0.1} \\
&LWR3k  
&\textbf{70\%} &\textbf{88.7$\pm$0.2} &\textbf{88.5$\pm$0.1}\\
&LWR5k  
&\textbf{70\%} &\textbf{88.7$\pm$0.1} &\textbf{88.5$\pm$0.1}\\

&LabelEmbed  
&- &88.3$\pm$0.1 &87.8$\pm$0.2 \\

&\cellcolor{gray!25}LwAL (Ours) 
&\cellcolor{gray!25}- &\cellcolor{gray!25}87.9$\pm$0.1 &\cellcolor{gray!25}87.5$\pm$0.0 \\

&\cellcolor{gray!25}LwAL10 (Ours) 
&\cellcolor{gray!25} -&\cellcolor{gray!25}87.6$\pm$0.2 &\cellcolor{gray!25}87.2$\pm$0.1 \\
&\cellcolor{gray!25}LwAL10+rpl (Ours) 
&\cellcolor{gray!25}- &\cellcolor{gray!25}88.2$\pm$0.1 &\cellcolor{gray!25}87.9$\pm$0.1 \\
	\hline
	\end{tabular}
\caption{All algorithms are trained with the same hyperparameter of learning rate ($=1e^{-4}$) over 10 epochs. \algo{LwAL} used 0 warmup steps and update frequency of once per step. Blank (--) indicates cases when the specific algorithm was unable to reach the reference \algo{STD} test accuracy. 
}
\label{tbl:text_dset_comparison}
\end{table*}

\subsection{Additional Results on Learning Speed and Test Performance}
\label{subsec:Additional Results on Learning Speed and Test Performance}
In addition to ``Percent Time/Epoch Reduced" and ``Best Test Accurarcy" in Table \ref{tbl:image_dset_resnet_performance_comparison}, we include average area under the accuracy curve (AUAC) in Table \ref{tbl:image_dset_resnet_performance_comparison_numbers}. This could be another useful metric to compare learning speed between algorithms, as larger area under the testing curve indicates a faster learning speed.

\subsection{Experiments with Text Dataset}
\label{subsec:Experiments with Text Dataset}
We also perform learning speed and test performance evaluations on text datasets, such as \dataset{IMDB reviews} \cite{imdb} and \dataset{Yelp Polarity Reviews} \cite{yelp}. Specifically, we first use BERT \cite{BERT} to extract a 768-dimensional representation of each text, and then use two Dense layers for predictions (one outputs 768 dimension, and another outputs number of classes). For \algo{StaticLabel}, we use BERT encodings of the word "negative" for class 0, and "positive" for class 1. We train all algorithms over 10 epochs, using ADAM with learning rate of 1e-4, and the rest of the training hyperparameters are the same discussed in the main text. The results are presented in Table \ref{tbl:text_dset_comparison}.

\subsection{Effects of Warmup Steps on LwAL}
\label{subsec:Effects of Warmup Steps on LwAL}
As discussed by \citet{LWR}, we experiment our \algo{LwAL} with some initial warmup steps $w$ to see if it can provide a better initial label separation and hence a better test performance. We experiment this with EfficientNetB0 backbone and report the results in Table \ref{tbl:warmup_steps_exp}. We find that using a few warmup steps can sometimes boost the test accuracy by a few percentage points. However, since this is not a consistent gain, we only presented results using $w=0$ in the main paper. In practice, this is a tuneable hyperparameter to further improve performance.

\begin{table*}[!th]
	\centering
	\begin{tabular}{|cc|ccc|}
		\hline
		\multicolumn{2}{|c|}{}
		&\multicolumn{3}{|c|}{Best Test Accuracy}\\
		& &$w=0$  &$w=2$ &$w=5$ \\
		\hline
\multirow{3}{*}{\dataset{MNIST}} 
&LwAL  
&99.3$\pm$0.1 &99.3$\pm$0.0 &99.3$\pm$0.1\\
&LwAL10  
&99.3$\pm$0.0 &99.3$\pm$0.1 &99.3$\pm$0.0\\
&LwAL10+rpl  
&99.3$\pm$0.0 &99.4$\pm$0.0 &99.3$\pm$0.1\\
	\hline\hline
\multirow{3}{*}{\dataset{Fashion MNIST}} 
&LwAL  
&93.0$\pm$0.2 &93.0$\pm$0.1 &93.2$\pm$0.0 \\
&LwAL10  
&92.7$\pm$0.2 &92.7$\pm$0.1 &92.8$\pm$0.1 \\
&LwAL10+rpl  
&92.8$\pm$0.2 &93.1$\pm$0.1 &93.0$\pm$0.2 \\
	\hline\hline
\multirow{3}{*}{\dataset{CIFAR10}} 
&LwAL  
&76.7$\pm$0.4 &76.8$\pm$0.2 &76.9$\pm$0.5 \\
&LwAL10  
&76.2$\pm$0.2 &76.1$\pm$0.8 &76.2$\pm$0.4 \\
&LwAL10+rpl  
&77.9$\pm$0.5 &78.3$\pm$0.2 &78.2$\pm$0.1\\
	\hline\hline
\multirow{3}{*}{\dataset{CIFAR100}}
&LwAL  
&43.2$\pm$0.2 &42.5$\pm$0.1 &42.5$\pm$0.2\\

&LwAL10  
&41.6$\pm$0.6 &41.7$\pm$0.4 &41.8$\pm$0.7\\

&LwAL10+rpl  
&42.2$\pm$0.5 &42.6$\pm$0.2 &42.3$\pm$0.4\\
	\hline\hline
\multirow{3}{*}{\dataset{FOOD101}} 
&LwAL  
&22.0$\pm$0.6 &22.1$\pm$0.1 &22.0$\pm$0.1\\

&LwAL10  
&20.5$\pm$0.5 &20.4$\pm$0.1 &20.3$\pm$0.1\\

&LwAL10+rpl  
&20.9$\pm$0.2 &20.9$\pm$0.1 &20.8$\pm$0.4\\
	\hline
	\end{tabular}
\caption{\algo{LwAL} warmup steps experiment with EfficientNetB0 backbone.}
\label{tbl:warmup_steps_exp}
\end{table*}

\begin{table*}[!th]
	\centering
	\begin{tabular}{|cc|ccc|ccc|}
		\hline
		\multicolumn{2}{|c|}{}
		&\multicolumn{3}{|c|}{Avg. AUAC}
		&\multicolumn{3}{|c|}{Best Test Accuracy}\\
      &&ResNet50  &EfficienNetB0 &DenseNet121  &ResNet50 &EfficienNetB0 &DenseNet121 \\
		\hline
\multirow{9}{*}{\rotatebox[origin=c]{90}{\dataset{MNIST}}} 
&One-hot (STD)      
&99.0$\pm$0.1 &98.3$\pm$0.1 &99.0$\pm$0.1 &99.1$\pm$0.1 &99.4$\pm$0.0 &99.1$\pm$0.1 \\
&StaticLabel  
&N/A    &N/A   	&N/A    &N/A    &N/A  &N/A   \\
&LWR2k  
&98.7$\pm$0.1 &\textbf{99.3$\pm$0.1} &99.0$\pm$0.1 &99.2$\pm$0.0 &99.4$\pm$0.1 &99.3$\pm$0.1 \\
&LWR3k  
&98.8$\pm$0.1 &99.2$\pm$0.0 &99.0$\pm$0.1 &99.1$\pm$0.1 &\textbf{99.5$\pm$0.1} &99.2$\pm$0.1 \\
&LWR5k  
&98.9$\pm$0.0 &99.1$\pm$0.1 &99.0$\pm$0.1 &99.2$\pm$0.1 &99.4$\pm$0.1 &99.2$\pm$0.1 \\

&LabelEmbed  
&99.0$\pm$0.1 &99.2$\pm$0.0 &\textbf{99.1$\pm$0.1} &99.2$\pm$0.1 &99.4$\pm$0.0 &99.4$\pm$0.0 \\

&\cellcolor{gray!25}LwAL (Ours) 
&\cellcolor{gray!25}98.9$\pm$0.0 &\cellcolor{gray!25}98.3$\pm$0.1 &\cellcolor{gray!25}98.9$\pm$0.0 &\cellcolor{gray!25}99.2$\pm$0.1 &\cellcolor{gray!25}99.3$\pm$0.1 &\cellcolor{gray!25}99.2$\pm$0.0\\

&\cellcolor{gray!25}LwAL10 (Ours) 
&\cellcolor{gray!25}98.9$\pm$0.0 &\cellcolor{gray!25}98.4$\pm$0.1 &\cellcolor{gray!25}98.9$\pm$0.0 &\cellcolor{gray!25}\textbf{99.3$\pm$0.1} &\cellcolor{gray!25}99.3$\pm$0.0 &\cellcolor{gray!25}99.1$\pm$0.0\\
&\cellcolor{gray!25}LwAL10+rpl (Ours) 
&\cellcolor{gray!25}\textbf{99.1$\pm$0.0} &\cellcolor{gray!25}98.4$\pm$0.0 &\cellcolor{gray!25}\textbf{99.1$\pm$0.0} &\cellcolor{gray!25}\textbf{99.3$\pm$0.1} &\cellcolor{gray!25}99.3$\pm$0.0 &\cellcolor{gray!25}\textbf{99.4$\pm$0.0}\\
	\hline\hline
\multirow{9}{*}{\rotatebox[origin=c]{90}{\dataset{Fashion MNIST}}} 
&One-hot (STD)      
&91.7$\pm$0.0 &92.1$\pm$0.1 &91.8$\pm$0.2 &92.3$\pm$0.2 &93.1$\pm$0.2 &92.4$\pm$0.3 \\

&StaticLabel  
&91.1$\pm$0.2 &91.5$\pm$0.1 &84.2$\pm$0.1 &92.8$\pm$0.1 &93.0$\pm$0.1 &92.6$\pm$0.2 \\

&LWR2k  
&91.3$\pm$0.3 &92.3$\pm$0.3 &91.7$\pm$0.3 &92.1$\pm$0.0 &93.3$\pm$0.3 &92.2$\pm$0.4 \\

&LWR3k  
&91.3$\pm$0.4 &\textbf{92.4$\pm$0.1} &91.7$\pm$0.3 &92.1$\pm$0.0 &\textbf{93.4$\pm$0.1} &92.3$\pm$0.4 \\

&LWR5k  
&91.5$\pm$0.2 &92.1$\pm$0.1 &91.7$\pm$0.2 &92.3$\pm$0.0 &\textbf{93.4$\pm$0.1} &92.5$\pm$0.2 \\

&LabelEmbed  
&\textbf{92.0$\pm$0.3} &92.1$\pm$0.2 &91.7$\pm$0.3 &92.7$\pm$0.4 &93.1$\pm$0.2 &92.9$\pm$0.1\\

&\cellcolor{gray!25}LwAL (Ours) 
&\cellcolor{gray!25}91.8$\pm$0.1 &\cellcolor{gray!25}92.0$\pm$0.1 &\cellcolor{gray!25}91.7$\pm$0.1 &\cellcolor{gray!25}\textbf{92.9$\pm$0.1} &\cellcolor{gray!25}93.0$\pm$0.2 &\cellcolor{gray!25}92.4$\pm$0.0\\

&\cellcolor{gray!25}LwAL10 (Ours) 
&\cellcolor{gray!25}91.6$\pm$0.0 &\cellcolor{gray!25}91.6$\pm$0.1 &\cellcolor{gray!25}91.9$\pm$0.1 &\cellcolor{gray!25}92.3$\pm$0.0 &\cellcolor{gray!25}92.7$\pm$0.2 &\cellcolor{gray!25}92.6$\pm$0.2\\

&\cellcolor{gray!25}LwAL10+rpl (Ours) 
&\cellcolor{gray!25}91.7$\pm$0.1 &\cellcolor{gray!25}91.5$\pm$0.1 &\cellcolor{gray!25}\textbf{92.1$\pm$0.1} &\cellcolor{gray!25}92.7$\pm$0.2 &\cellcolor{gray!25}92.8$\pm$0.2 &\cellcolor{gray!25}\textbf{93.0$\pm$0.2}\\
	\hline\hline

\multirow{9}{*}{\rotatebox[origin=c]{90}{\dataset{CIFAR10}}} 
&One-hot (STD)      
&70.8$\pm$0.4 &72.8$\pm$0.2 &76.8$\pm$0.3 &73.3$\pm$0.5 &75.9$\pm$0.4 &78.8$\pm$0.5 \\

&StaticLabel  
&52.0$\pm$1.5 &67.5$\pm$1.1 &49.1$\pm$0.6 &74.0$\pm$0.7 &75.7$\pm$0.5 &77.7$\pm$0.3 \\

&LWR2k  
&64.1$\pm$2.5 &71.9$\pm$0.1 &72.1$\pm$1.7 &67.8$\pm$1.1 &74.7$\pm$0.3 &74.1$\pm$0.7 \\

&LWR3k  
&65.1$\pm$1.7 &72.3$\pm$0.2 &73.1$\pm$1.2 &69.3$\pm$0.8 &75.3$\pm$0.1 &75.6$\pm$0.9 \\

&LWR5k  
&67.8$\pm$1.0 &72.7$\pm$0.3 &75.0$\pm$0.9 &69.9$\pm$1.1 &76.3$\pm$0.4 &76.9$\pm$0.7 \\

&LabelEmbed  
&68.3$\pm$0.8 &72.4$\pm$0.1 &76.2$\pm$0.5 &72.2$\pm$0.9 &76.7$\pm$0.3 &79.4$\pm$0.4\\

&\cellcolor{gray!25}LwAL (Ours) 
&\cellcolor{gray!25}70.8$\pm$0.2 &\cellcolor{gray!25}73.3$\pm$0.4 &\cellcolor{gray!25}76.7$\pm$0.2 &\cellcolor{gray!25}72.0$\pm$0.5 &\cellcolor{gray!25}76.7$\pm$0.4 &\cellcolor{gray!25}78.9$\pm$0.0 \\

&\cellcolor{gray!25}LwAL10 (Ours) 
&\cellcolor{gray!25}72.4$\pm$0.1 &\cellcolor{gray!25}72.5$\pm$0.1 &\cellcolor{gray!25}77.5$\pm$0.6 &\cellcolor{gray!25}73.9$\pm$0.0 &\cellcolor{gray!25}76.2$\pm$0.2 &\cellcolor{gray!25}79.2$\pm$0.4 \\

&\cellcolor{gray!25}LwAL10+rpl (Ours) 
&\cellcolor{gray!25}\textbf{73.4$\pm$0.1} &\cellcolor{gray!25}\textbf{74.4$\pm$0.4} &\cellcolor{gray!25}\textbf{78.0$\pm$0.5} &\cellcolor{gray!25}\textbf{76.0$\pm$0.4} &\cellcolor{gray!25}\textbf{77.9$\pm$0.5} &\cellcolor{gray!25}\textbf{80.5$\pm$0.3}\\
	\hline\hline

\multirow{9}{*}{\rotatebox[origin=c]{90}{\dataset{CIFAR100}}}
&One-hot (STD)      
&30.3$\pm$0.1 &35.4$\pm$0.4 &35.6$\pm$0.9 &37.4$\pm$0.6 &40.5$\pm$0.5 &44.6$\pm$0.8\\

&StaticLabel  
&6.1$\pm$0.2$^{*}$ &2.6$\pm$0.4 &2.8$\pm$0.0$^{*}$ &16.8$\pm$1.1$^{*}$ &5.9$\pm$0.6 &7.8$\pm$0.3$^{*}$\\

&LWR2k  
&27.1$\pm$0.4 &33.6$\pm$0.7 &32.0$\pm$0.5 &32.9$\pm$0.5 &38.1$\pm$0.5 &38.7$\pm$0.4\\

&LWR3k  
&27.1$\pm$0.2 &33.8$\pm$0.4 &32.0$\pm$0.6 &32.7$\pm$0.1 &38.1$\pm$0.4 &38.6$\pm$0.6\\

&LWR5k  
&27.4$\pm$0.1 &33.9$\pm$0.6 &32.2$\pm$0.5 &32.9$\pm$0.4 &38.1$\pm$0.7 &38.6$\pm$0.6\\

&LabelEmbed  
&25.9$\pm$0.5 &34.6$\pm$0.5 &32.0$\pm$1.4 &37.7$\pm$0.8 &41.0$\pm$0.5 &44.6$\pm$0.7 \\

&\cellcolor{gray!25}LwAL (Ours) 
&\cellcolor{gray!25}36.5$\pm$0.4 &\cellcolor{gray!25}\textbf{38.6$\pm$0.3} &\cellcolor{gray!25}42.6$\pm$0.1 &\cellcolor{gray!25}38.8$\pm$0.4 &\cellcolor{gray!25}\textbf{43.2$\pm$0.2} &\cellcolor{gray!25}46.8$\pm$0.3 \\

&\cellcolor{gray!25}LwAL10 (Ours) 
&\cellcolor{gray!25}37.3$\pm$0.2 &\cellcolor{gray!25}37.9$\pm$0.5 &\cellcolor{gray!25}\textbf{44.1$\pm$0.3} &\cellcolor{gray!25}39.3$\pm$0.2 &\cellcolor{gray!25}41.6$\pm$0.6 &\cellcolor{gray!25}47.5$\pm$0.4\\

&\cellcolor{gray!25}LwAL10+rpl (Ours) 
&\cellcolor{gray!25}\textbf{37.5$\pm$0.2} &\cellcolor{gray!25}38.4$\pm$0.4 &\cellcolor{gray!25}43.7$\pm$0.1 &\cellcolor{gray!25}\textbf{39.9$\pm$0.4} &\cellcolor{gray!25}42.2$\pm$0.5 &\cellcolor{gray!25}\textbf{48.0$\pm$0.0}\\
	\hline\hline

\multirow{9}{*}{\rotatebox[origin=c]{90}{\dataset{FOOD101}}} 
&One-hot (STD)      
&12.7$\pm$0.2 &16.3$\pm$0.4 &16.7$\pm$0.1 &16.3$\pm$0.3 &18.5$\pm$0.5 &20.6$\pm$0.0\\

&StaticLabel  
&1.3$\pm$0.1$^{*}$ &1.3$\pm$0.3 &2.2$\pm$0.1$^{*}$ &2.6$\pm$0.5$^{*}$ &2.0$\pm$0.8 &6.4$\pm$0.5$^{*}$\\

&LWR2k  
&10.7$\pm$0.1 &16.2$\pm$0.3 &14.8$\pm$0.2 &13.8$\pm$0.1 &18.0$\pm$0.3 &17.9$\pm$0.2 \\

&LWR3k  
&10.8$\pm$0.1 &16.2$\pm$0.3 &14.8$\pm$0.2 &13.9$\pm$0.1 &18.2$\pm$0.3 &18.0$\pm$0.3 \\

&LWR5k  
&10.9$\pm$0.0 &16.5$\pm$0.4 &14.9$\pm$0.2 &13.9$\pm$0.1 &18.5$\pm$0.4 &17.8$\pm$0.1\\

&LabelEmbed  
&9.7$\pm$0.2 &16.7$\pm$0.2 &15.6$\pm$0.3 &15.8$\pm$0.1 &19.8$\pm$0.3 &21.6$\pm$0.5 \\

&\cellcolor{gray!25}LwAL (Ours) 
&\cellcolor{gray!25}14.5$\pm$0.1 &\cellcolor{gray!25}\textbf{19.6$\pm$0.4} &\cellcolor{gray!25}18.1$\pm$1.0 &\cellcolor{gray!25}16.6$\pm$0.3 &\cellcolor{gray!25}\textbf{22.0$\pm$0.6} &\cellcolor{gray!25}21.1$\pm$0.1 \\

&\cellcolor{gray!25}LwAL10 (Ours) 
&\cellcolor{gray!25}15.8$\pm$0.1 &\cellcolor{gray!25}18.7$\pm$0.4 &\cellcolor{gray!25}19.7$\pm$0.2 &\cellcolor{gray!25}17.5$\pm$0.3 &\cellcolor{gray!25}20.5$\pm$0.5 &\cellcolor{gray!25}22.5$\pm$0.1\\

&\cellcolor{gray!25}LwAL10+rpl (Ours) 
&\cellcolor{gray!25}\textbf{16.0$\pm$0.2} &\cellcolor{gray!25}19.0$\pm$0.2 &\cellcolor{gray!25}\textbf{20.1$\pm$1.0} &\cellcolor{gray!25}\textbf{17.7$\pm$0.1} &\cellcolor{gray!25}20.9$\pm$0.2 &\cellcolor{gray!25}\textbf{22.9$\pm$0.1}\\
	\hline
	\end{tabular}
\caption{Learning accuracy and speed comparison between \algo{LwAL} and other baselines.
\algo{LwAL} is trained using 0 warmup steps and update frequency of once per step. 
Star (*) indicates the use of different learning rate ( $1e^{-3}$) due to failure of convergence. N/A for \dataset{MNIST} dataset using \algo{StaticLabel} indicates that the BERT representation of \dataset{MNIST} categories is not appropriate.}
\label{tbl:image_dset_resnet_performance_comparison_numbers}
\end{table*}

\end{document}